\documentclass{bmvc2k}
\usepackage{wrapfig,lipsum,booktabs}
\usepackage{amsmath}
\usepackage{graphicx}
\usepackage{amssymb}
\usepackage{color}
\usepackage{multirow}

\usepackage{tabu}
\usepackage{float}
\usepackage[font={color=bmv@captioncolor}]{subcaption}
\usepackage[font={color=bmv@captioncolor}]{caption}
\usepackage{xspace}
\usepackage[shortcuts]{extdash}

\newif\ifprintcomments
\title{OWL (Observe, Watch, Listen):\\ Audiovisual Temporal Context for Localizing Actions in Egocentric Videos}

\addauthor{Merey Ramazanova}{merey.ramazanova@kaust.edu.sa}{1}
\addauthor{Victor Escorcia}{victor.escorcia@kaust.edu.sa}{2}
\addauthor{Fabian Caba Heilbron}{caba@adobe.com}{3}
\addauthor{Chen Zhao}{chen.zhao@kaust.edu.sa}{1}
\addauthor{Bernard Ghanem}{bernard.ghanem@kaust.edu.sa}{1}

\addinstitution{
 King Abdullah University of Science and Technology\\
}
\addinstitution{
 Samsung AI Center Cambridge \\
}
\addinstitution{
 Adobe Research\\
}
\runninghead{Ramazanova et al.}{OWL (Observe, Watch, Listen)}

\def\eg{\emph{e.g}\bmvaOneDot}

\def\etal{\emph{et al}\bmvaOneDot}
\def\vs{\emph{vs}\bmvaOneDot}
\newcommand{\cf}{{\it cf.\ }}
\begin{document}

\maketitle

%%%%%%%%% ABSTRACT
\begin{abstract}
% Temporal action localization (TAL) is an important task extensively explored and improved for third-person videos in recent years. Recent efforts have been made to perform fine-grained temporal localization on first-person videos. 

%Egocentric videos capture sequences of human activities, and can provide rich multimodal signals. However, 

%Egocentrics videos capture human activities from a first person perspective. These videos are often densely recorded and unedited. Temporal action  localization has been studied mostly on third-person videos in which the audio modality has not been widely exploded. In this work...
% In egocentric videos, actions are dense and capture sequences of human activities.
Egocentric videos capture sequences of human activities from a first-person perspective and can provide rich multimodal signals. However, most current localization methods use third-person videos and only incorporate visual information. In this work, we take a deep look into the effectiveness of audiovisual context in detecting actions in egocentric videos and introduce a simple-yet-effective approach via Observing, Watching, and Listening (OWL). OWL leverages audiovisual information and context for egocentric temporal action localization (TAL).
We validate our approach in two large-scale datasets, EPIC-Kitchens, and HOMAGE. Extensive experiments demonstrate the relevance of the audiovisual temporal context. Namely, we boost the localization performance (mAP) over visual-only models by +2.23\% and +3.35\% in the above datasets.
%We perform extensive experiments on EPIC-Kitchens, where we achieve competitive performance and boose the baseline performace of $7.06\%$ to $9.29\%$ . We also make the first attempt to explore HOMAGE, a more versatile dataset for egocentric TAL. We improve the baseline performance from $6.16\%$ to $9.51\%$ and verify the effectiveness of our method beyond cooking-related videos. 

\end{abstract}

%%%%%%%%% BODY TEXT
\section{Introduction}
\label{sec:intro}
%CHALLENGES OF TAL IN EGO -> IMPORTANCE OF MULTI-MODAL CONTEXT IN TAL 
Egocentric videos capture the world using wearable cameras. Arguably, in these videos, localizing actions in time is top of mind~\cite{damen2018scaling}. In doing so, we could enable world-changing applications such as an episodic memory AI assistant for health monitoring.
% ranging from generating consumer highlights to building artificial episodic memory for health monitoring.
Localizing and recognizing human actions in egocentric video imposes several challenges. Due to the capture nature, videos tend to be long and highly unconstrained w.r.t.~the activities occurring on the stream. Given that the capture happens through a camera mounted on a person's head, challenging conditions such as undesired camera motions, occlusions, and poor quality video make the problem of localizing and recognizing actions a complex task. Additionally, existing egocentric datasets, \eg~\cite{damen2021rescaling}, focus on localizing atomic actions that happen densely across long videos. Consequently, the performance of egocentric TAL lags far behind compared to that in the third-person setting~\cite{nawhal2021activity}. Given such complexity, analyzing the relationships of actions and looking beyond visual cues is essential in an egocentric scene.

\begin{figure}[t!]
    \captionsetup[subfloat]{font={scriptsize, color=bmv@captioncolor},labelfont=scriptsize}
    \centering
    \subfloat[]{%
      \includegraphics[clip,width=\columnwidth]{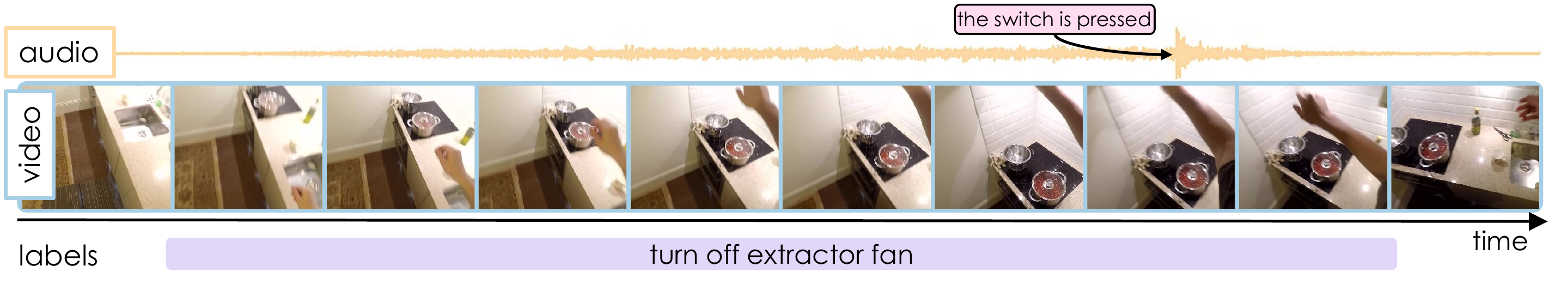}%
        \label{subfig:pull_audio}
    }

    \subfloat[]{%
      \includegraphics[clip,width=\columnwidth]{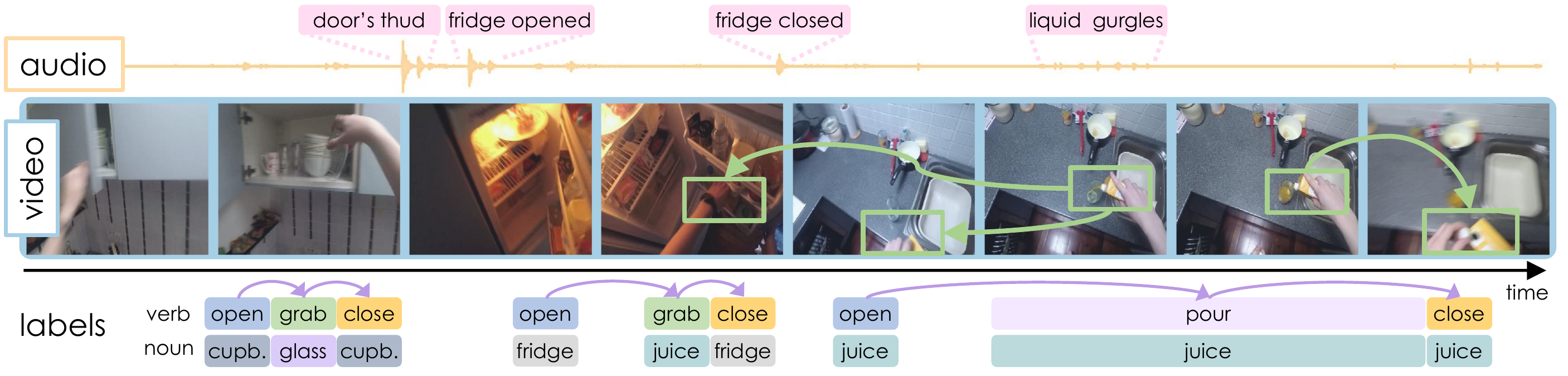}%
        \label{subfig:pull_context}
    }

    \caption{
    {\bf Audiovisual temporal context is an important cue for the temporal localization of actions in egocentric unedited videos}. In video \textbf{(a)}, the action, \textit{turning off the extractor fan}, is more evident when observing the interplay between audio and visual streams. The fan is invisible but the interruption of the humming noise in the audio signal provides context to the movement of the hand in the visual domain.
    In  video \textbf{(b)}, the recorder is preparing a glass of juice . The \textcolor{green}{green drawn boxes} spatially localize the juicebox. Knowing the content of the box in action \textit{pour juice} could help in predicting ambiguous actions \textit{open juice}, \textit{close juice}, and \textit{grab juice} (\textcolor{green}{green arrows}). By following the \textcolor{violet}{violet arrows} in the annotations, we can see the pattern of how people interact with kitchen items.
    % \fch{While this figure is pretty descriptive, can we think of another illustration that conveys our contribution in a single shot? There is a lot of distractor elements that force the reader to: (1) read the entire caption, (2) read the intro, (3) understand about attributes of the EpicKitchen dataset (verb, noun). Also, the green boxes might confuse the reader: one might think we are doing spatial localizations.}
    }
    \label{figure:teaser}
    % \vspace{-3mm}
\end{figure}

%SPECIFICS OF EGO VIDEOS -> GOOD AUDIO 
% Despite its challenges, there are particular properties of the current egocentric datasets~\cite{damen2021rescaling,Grauman_2022_CVPR,rai2021home} to benefit TAL.
Despite its challenges, there are particular properties of the current egocentric datasets~\cite{damen2021rescaling,rai2021home} to benefit TAL.
Since the videos are \textit{unedited} and \textit{continuous}, the audio stream is synchronized with the visual stream, capturing the sounds and appearance of what is happening in the video at the moment. This is different from videos in traditional datasets that are curated from online video platforms like YouTube. In such datasets and due to the editing, audio might not correspond to the original sounds present in the scene. We argue that audio, in egocentric video, plays an important role in assisting visual models to localize human actions. For example, looking at Fig.~\ref{subfig:pull_audio}, we notice a person reaching for something in a kitchen. Because of the camera view, we cannot see the object they are interacting with. Can we guess what exactly are they doing? By observing the lighting and the location (above the stove), we could imagine the interaction with the fan. But how can we discern if the fan was turned off or on? By hearing the sounds from the scene, you would not doubt that the person is \textit{`turning off the extraction fan'}. The fan's distinctive humming noise and its disappearance indicate the action happening and its precise temporal endpoints.
%AV CONTEXT

Using temporal context has been proven to be effective for both action recognition, and localization~\cite{cartas2021modeling,ng2019human,wu2019long,zhang2021temporal,kazakos2021little,xu2020g,zeng2019graph}. Temporal context might be even more informative in egocentric videos. For instance, at being unedited and continuous, actions unfold, with a more often than not, predictable sequence~\cite{furnari2019would,Girdhar_2021_ICCV,kazakos2021little}. 
To illustrate how context can be helpful to localize egocentric actions, we present a toy example in Fig.~\ref{subfig:pull_context}. Looking at the sequence holistically, the scenario is clear: the recorder is preparing a glass of juice. If we look at each shot separately (imitating a neural network classifying a trimmed clip), we could probably struggle to recognize some actions. It is unclear that the box, which the recorder is grabbing from the fridge, then opening and closing, contains juice. When we see some orange liquid (and hear) pouring from it, we can guess it must be orange juice. The instances \textit{`grab juice'} and \textit{`pour juice'} are almost five seconds away, but still are informative to each other. Moreover, by leveraging context, we can decode the sequential patterns of actions in cooking activities. We argue that audiovisual context provide priors to better localize actions.
% These intuitions motivate us to investigate the role of audiovisual context in \textit{egocentric} TAL. 
% dai2017temporal
%  Compared to the existing literature in third-person temporal action localization (TAL)~\cite{lin2019bmn,xu2020g,bai2020boundary,zhao2021video}, and despite its importance, TAL in egocentric videos remains relatively underexplored. 
%  

We propose \textbf{OWL (Observe, Watch, Listen)}, a simple-yet-effective transformer-based architecture that leverages audiovisual context to localize actions in egocentric videos. 
We do a methodical analysis to verify the importance of audiovisual context in egocentric videos.
% /merey{To develop OWL, we performed a detailed exploration of plausible design choices.} We also explore what part in the feature space would be best to fuse the audiovisual information (Supplementary Material, Tab.~3).
First, we study which components of the action localization pipeline would benefit from audio cues (Sec.\ref{sec:experiments}, Tab.~\ref{tab:bmnxclassifier}).  Furthermore, we analyze what temporal neighborhood provides the richer context (Sec.\ref{sec:experiments}, Tab.\ref{tab:context_size}). Finally, we analyze how visually occluded instances largely benefit from context in egocentric videos (Sec.\ref{sec:experiments}, Tab.\ref{tab:occlusion}). 
OWL uses self-attention to encode context within each modality and cross-attention to capture relevant context across modalities.
Our experiments on EPIC-Kitchens-100 (EK100)~\cite{damen2021rescaling} and HOMAGE~\cite{rai2021home} validate that OWL effectively encodes audiovisual context for egocentric TAL and significantly improves over proposed audiovisual baselines.

\noindent\textbf{Contributions.} 
\textbf{(1)} We propose a transformer-based method for egocentric action localization by effectively fusing audiovisual context (Sec.~\ref{sec:OWL}).
\textbf{(2)} We conduct extensive experiments on EK100 and HOMAGE in Sec.~\ref{sec:quantitative}, and achieve competitive results.
\textbf{(3)} We conduct a thorough analysis that validates our hypothesis and findings about audiovisual context for action localization in egocentric videos (Sec.~\ref{sec:experiments-analysis}).

% \textbf{(1)} We provide a thorough analysis on the effectiveness of the audio modality to boost localization performance. 
% We design multiple alternatives to fuse audio into traditional action localization pipelines (Sec.~\ref{subsec:fusion}).
% \textbf{(2)} We propose, OWL, a transformer-based method for egocentric action localization, that effectively fuses audiovisual context (Sec.~\ref{sec:OWL}).

% \merey{\textbf{(4)} We analyze the effectiveness of OWL in the occluded environments to validate the importance of audio and context in Sec.~\ref{sec:occlusion}}.
% We compare the performance of the different fusion alternatives and OWL. All in all, we argue that using audiovisual signals  egocentric action localization.

\section{Related Work}
\label{sec:related}

% Victor just reordered

\noindent\textbf{Audiovisual learning.}
Video and audio are common modality choices for a multi-modal learning scenario in video understanding. Deep learning facilitates audiovisual learning as it enables learning per-modality hierarchical representations~\cite{ramachandram2017deep}, which are more optimal than designing hand-crafted features. Recent works provide us with more sophisticated solutions where the learned modality representations are being fused implicitly by the network and are optimized for the downstream task, such as ~\cite{ephrat2018looking,alcazar2021maas,xiao2020audiovisual,wang2020makes,nagrani2021attention,kazakos2019epic,kazakos2021little}. 
While several works discussed the audiovisual scenario for the action recognition task~\cite{xiao2020audiovisual}, incorporating audio for TAL is not a widely researched area.~\cite{tian2018audio} proposes a new task of audiovisual event localization that aims at predicting the event class from a 10-second clip.~\cite{bagchi2021hear} studies multi-modal fusion approaches for audiovisual localization but ablates it on third-person datasets. Compared to them, we design our method for long, diverse egocentric videos. We are particularly motivated by~\cite{kazakos2019epic}, who emphasized the advantage of using egocentric unedited videos for applying audiovisual learning in action recognition. To the best of our knowledge, we are the first work that analyzes this advantage in egocentric TAL.
% \fch{This is a bit contradictory to our intro arguments of audio being more (and probably only) helpful (compared to third-person videos) to localize actions in egocentric videos. Instead of saying that the method "validates", can we point out some limitations?}

\noindent\textbf{Temporal action localization (TAL).}
Given an untrimmed video, TAL models aim to detect the boundaries and classes of all actions happening inside the video. Recent work can be categorized into separate-stage and combined-stage methods. The separate-stage methods generate a set of class-agnostic proposals (generation) first and then use a separate classifier to assign an action class to each proposal~\cite{bai2020boundary,caba2017scc,escorcia2016daps,lin2018bsn,lin2019bmn,xu2020g}. Most of the existing separate-stage methods focus on generating better proposals and rely on global video classification models and dataset statistics to classify them. Combined-stage solutions perform action localization in one unified pipeline by optimizing for both tasks simultaneously ~\cite{xu2017r,liu2020progressive,zhao2021video,nawhal2021activity}. In this paper we follow the separate-stage approach.

\noindent\textbf{Egocentric (unedited) videos.}
TAL has been extensively studied for third-person and mostly edited videos (typically, from consumer media platforms like YouTube and movies)~\cite{caba2015activitynet,jiang2014thumos,zhao2019hacs}. The appearance of new large-scale egocentric datasets~\cite{damen2021rescaling,rai2021home} opened up a unique opportunity for researchers to study human actions in unedited videos. The annotations for action localization in most common (third-person) benchmarks are relatively sparse, with a low variation in assigned classes per video (ActivityNet~\cite{caba2015activitynet} has on average $1.5$ instances and $1.0$ class per video, in THUMOS14 these numbers are $15.4$ and $1.1$, respectively). That makes it possible to condition the localized action class by gathering visual cues at the video-level. This paradigm is not suitable for more dense and diverse datasets. For instance, EK100 has on average $128.5$ instances and $53.2$ classes per video. That said, assigning proposals with a single video-level class would yield pretty poor localization results. To address the densely annotated videos on EK100, Damen \etal introduce a baseline separate-stage approach using BMN~\cite{lin2019bmn} proposals and SlowFast~\cite{feichtenhofer2019slowfast} classification. \cite{nawhal2021activity} proposes a combined-stage method (AGT) that leverage graph-based and transformer-based architectures to localize and classify actions jointly. Note that these approaches do not explicitly (or implicitly) model temporal context or leverage the egocentric audio streams. Our work lies in the separate-stage group; thus, to design OWL, we thoroughly investigate effective multi-modal and contextualized classifiers to assign each proposal an action class.

\noindent\textbf{Temporal context in action localization.}
The importance of temporal context has been a long-standing aspect in action localization~\cite{dai2017temporal,alwassel2018action,wu2019long,zeng2019graph,xu2020g,qing2021temporal}. Some works~\cite{zeng2019graph,xu2020g} propose graph-based methods, where they define proposals and snippets as graph nodes and perform graph convolutions for the information exchange.
Our approach is closer to recent work that leverage the Transformer architecture~\cite{liu2021end,nawhal2021activity,sridhar2021class}. Due to the rising popularity of transformers for vision tasks~\cite{dosovitskiy2020image,carion2020end,arnab2021vivit}, a few works \cite{liu2021end,nawhal2021activity,sridhar2021class} extended the transformer building blocks to the inner working of TAL as a way to infuse temporal context between proposals.
In contrast to prior art, our work considers the interplay of multiple modalities, visual and audio, while also modeling the surrounding context of an action. By putting \emph{audiovisual context} at the fore front, architectural differences arises in comparison to existing transformer-based approaches.

\section{Methodology}
\label{sec:method}

% \input{figures/fusion}

\begin{figure}
\captionsetup[subfloat]{font={tiny, color=bmv@captioncolor},labelfont=scriptsize}
\centering

\begin{subfigure}[b]{.42\textwidth}
%   \centering
  \includegraphics[width=\linewidth]{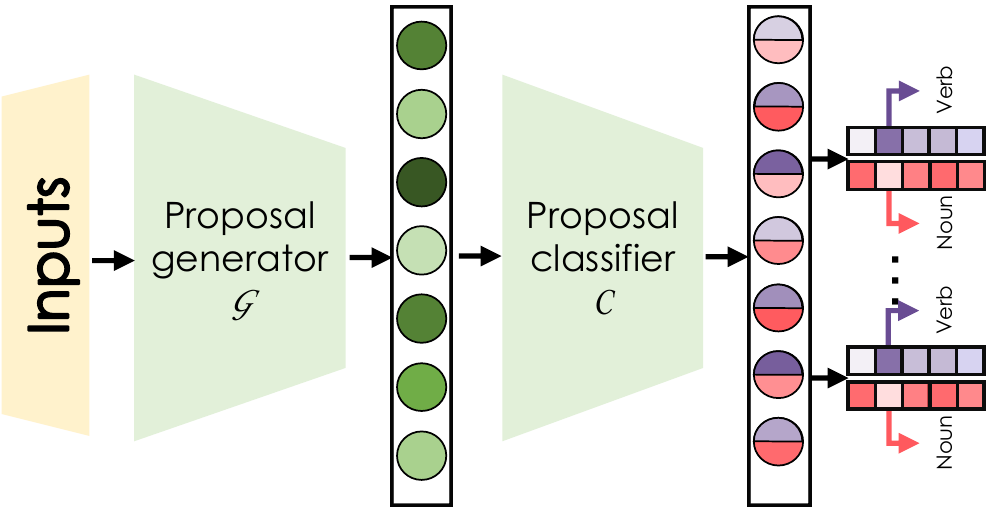}
  \caption{}
  \label{subfig:block}
\end{subfigure}%
\hfill
\begin{subfigure}[b]{.54\textwidth}
%   \centering
  \includegraphics[width=\linewidth]{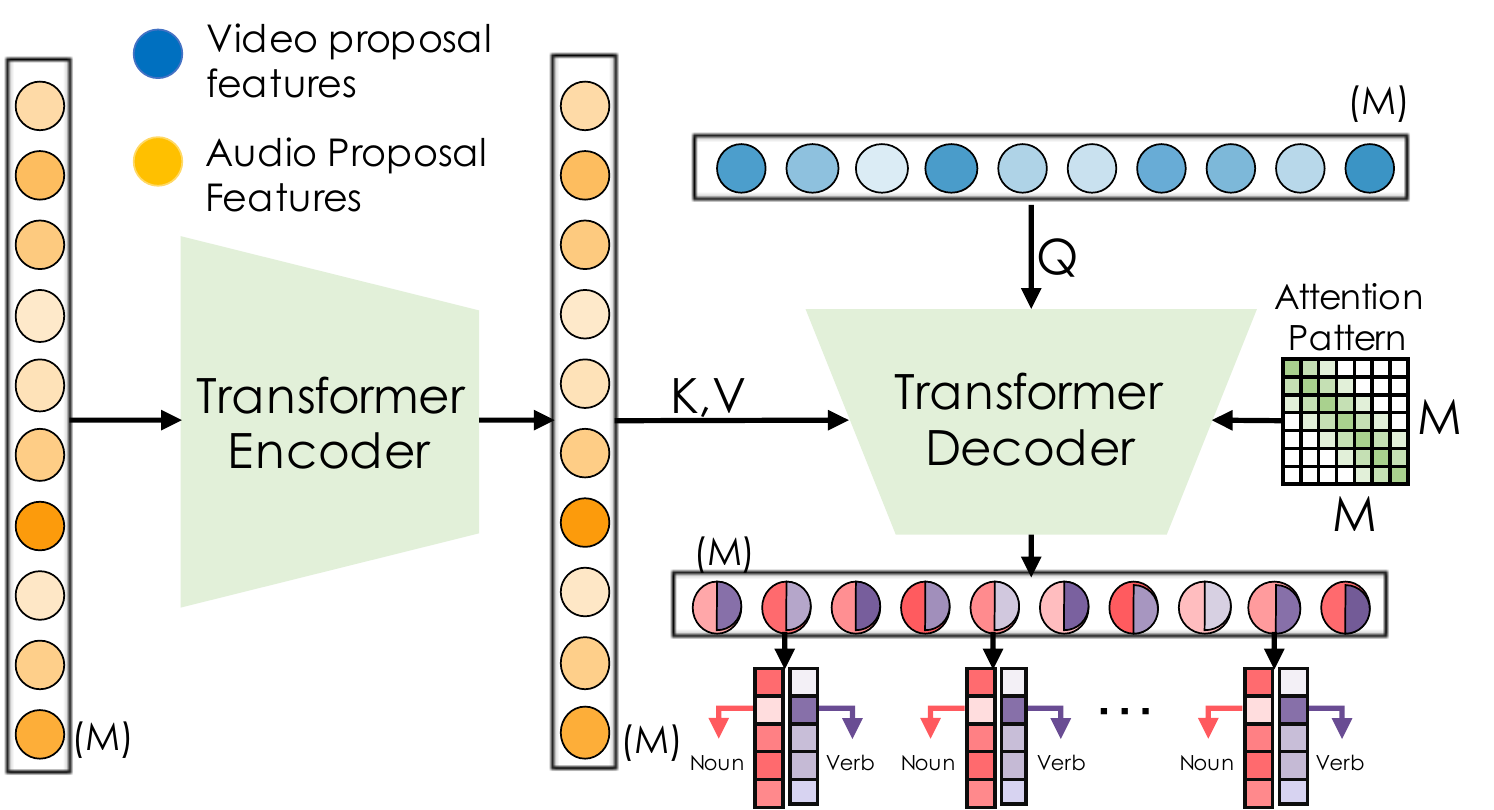}
  \caption{}
  \label{subfig:context}
\end{subfigure}
\caption{ \textbf{(a) Separate-stage pipeline for TAL.} Given a sequence of snippet features, $\mathcal{G}$ produces class\=/agnostic action proposals with start/end timestamps. Then, $\mathcal{C}$ takes a set of proposal features and produces classification labels for each proposal. \textbf{(b) OWL:} We input the auditory sequence (\textcolor{yellow}{yellow}) into the encoder and the visual sequence (\textcolor{blue}{blue}) into the decoder. $K, V$, and $Q$ refer to the components of multi-head attention as in \cite{vaswani2017attention}. The encoder and decoder first perform self-attention to enrich the intra-modal representations. Then, the decoder performs multi-head cross-attention. The amount of context $W$ (the \textcolor{green}{green} band on the attention pattern), within which self-attention and cross-attention act,  can be controlled by the attention mask of size $M \times M$. $M$ is the input sequences size (number of proposals).}
\label{figure:block_context}
\end{figure}

% \subsection{Separate-stage framework}
Given a sequence of video frames $V=\{I_t\}_{t=1}^T$ , the task of TAL is to predict  a set of  segments
%$\Psi  = \left \{ \psi_n  \right \}_{n=1}^{N}$
$\Psi  = \left \{ \tau_n, s_n, y_n  \right \}_{n=1}^{N}$
with start/end timestamps $\tau_n$, confidence score $s_n$ and action class labels $y_n$. In our work, we consider both the visual and audio modalities of the video sequence.  We first encode either modality  into snippet\=/level features $\mathbf{x} \in \mathbb{R}^{D\times L}$\ \cite{escorcia2016daps, xu2020g}, where $L$ is the number of encoded snippets and $D$ is the channel dimension. The feature encoder usually adopts the pre-trained backbone of an action recognition model, such as ~\cite{TSN2016ECCV, feichtenhofer2019slowfast}. 
% In EK100 action labels are determined by the combination of noun and verb labels, which are predicted for each segment.
Our approach follows a \textbf{ separate\=/stage pipeline}, where \textit{Proposal generator} $\mathcal{G}$ generates class-agnostics proposals $\Psi_{\mathcal{G}} = \left \{ \tau_n, s_n \right \}_{n=1}^{N}$, and then the \textit{Proposal classifier} $\mathcal{C}$ assigns a class label $y_n$ to them (including background class), as shown in Fig.~\ref{subfig:block}.
% To see a more formal definition of $\mathcal{G}$ and $\mathcal{C}$, see the supplementary.
% \fch{Could we indicate in Figure 2a, $\mathcal{G}$ and $\mathcal{C}$ in the pipeline blocks?}

% \fch{is attention mask the same as attention window? if so, then here we should replace $M$ by $W$. Note that we also use $M$ for the number of proposals generated by $\mathcal{G}$}

\vspace{3pt}\noindent\textbf{Observe, watch, and listen.}
% \subsection{Observe, watch, and listen.}
\label{sec:OWL}
 We propose OWL (\textbf{O}bserve, \textbf{W}atch, \textbf{L}isten), a transformer based model\ \cite{vaswani2017attention}, to leverage multi\=/modal context in TAL. It uses an encoder composed of a self\=/attention module to encode the audio features, and a decoder composed of a self\=/attention and a cross\=/attention modules to encode the visual features and to fuse both modalities (Fig.~\ref{subfig:context}). Besides \textbf{watching} the visual signal and \textbf{listening} to the audio signal, our OWL is also able to \textbf{observe} each proposal in the context of its neighbours proposals. We model the visual and audio proposal\=/level features $\mathbf{z}^v$ and $\mathbf{z}^a$ as the input tokens for the transformer.  We use the superscripts ${v}$ and ${a}$ for the visual and audio modalities, respectively.  
%   Both transformer encoder and decoder contain a self-attention module, where the features of all proposals for either modality are input tokens that generate Query(Q), Key(K), and Value(V) via different MLPs.

\noindent\textbf{Positional encodings.}
As transformer operations are permutation invariant, we use positional encodings to preserve the temporal relationship of the proposals. We encode the relative proposal start time and its absolute duration. The relative start time $p_{r}$ incorporates the position of an action in the video and the temporal order of actions. By encoding the absolute duration $p_{d}$, we inject the temporal information that is lost after pooling. Specifically,
$p_{d} = t_{\textrm{e}} - t_{\textrm{s}}$ and 
$p_{r} = \frac{t_{\textrm{s}}}{T}$, where $t_{\textrm{e}}$ and $t_{\textrm{e}}$ are the proposal's predicted start and end times, respectively. 
We pass $p_{r}$ and $p_{d}$ to a fully-connected (FC) layer to generate the positional encoding $\mathbf{p} \in \mathbb{R}^{D^e}$\ \cite{Escorcia2019TemporalLO}. $\mathbf{p}$ is concatenated to $\mathbf{z}^v$ and $\mathbf{z}^a$ and passed to the transformer encoder.
\label{subsec:context}

\noindent\textbf{Intra-modal \& inter-modal context.} For each token of either modality, the self-attention module \textbf{observes its relevant intra-modal context}, correlating other proposals to enhance its feature representation.  After self-attention, we obtain enhanced representations $\mathbf{z}_e^v$ and $\mathbf{z}_e^a$ for each proposal. The transformer decoder fuses both modalities. It contains a cross-attention module, which takes   $\mathbf{z}_e^v$ and $\mathbf{z}_e^a$ as input tokens. The visual modality tokens are used as queries $Q$, and audio modality tokens are used keys $K$ and values $V$ (Fig. \ref{subfig:context}). Recall that attention mechanism transforms $Q, K, V$ as 
\begin{align}
  Attention(Q,K,V) = softmax \left(\frac{QK^T}{\sqrt{D}}\right)V.
\end{align}
Hereby, the audio features are linearly combined based on the similarities between video and audio proposal\=/level features. The resulting features are enriched by \textbf{observing the inter-modal context} from neighboring proposals.
%audio and video features.
Theoretically, we can correlate all $M$ proposals in a video, but to study how much context is needed, we restrict the self-attention and the cross-attention to attend only to the proposals within a temporal neighborhood $W$ (inspired by~\cite{beltagy2020longformer}). As shown on Fig.~\ref{subfig:context}, each proposal can attend to only $\frac{W}{2}$ tokens from each side.
% \subsection{Training and inference.}
\vspace{3pt}

\noindent\textbf{Training and inference.} We generate classification scores based on the enriched proposal\=/level features produced by OWL. We train $\mathcal{C}$ using standard cross-entropy loss. During inference, we multiply the scores of each noun and verb pairs to generate the action scores.
% with the following joint loss:
% \begin{align}
%     \mathcal{L} = \lambda^{\textrm{noun}}\mathcal{L}^{\textrm{noun}} + \lambda^{\textrm{verb}} \mathcal{L}^{\textrm{verb}},
% \end{align} 

%%%% TO BE MOVED TO THE SUPP START  %%%%

%%%% TO BE MOVED TO THE SUPP END  %%%%
%~\cite{vaswani2017attention}.

\section{Experiments}
\label{sec:experiments}
\subsection{Dataset}
We evaluate our proposed method on two large-scale egocentric video datasets. EK100~\cite{damen2021rescaling} contains 700 unscripted videos of people performing their daily kitchen routines. It has, on average, 129 annotated instances per video, which make it significantly harder to perform TAL compared to the established benchmarks~\cite{jiang2014thumos,caba2015activitynet, zhao2019hacs}. Around 28\% of actions overlap, and each annotated instance is composed of a verb and a noun pair describing an action performed with an object. Overall, there are 300 noun and 97 verb classes.

 HOMAGE \cite{rai2021home} is a multi-view action dataset with audiovisual synchronized video data, containing a diverse set of daily activities. It has, on average, 15 instances per video, and  90$\%$ of the scenes in HOMAGE have the egocentric view. The action annotations for HOMAGE are not decomposed into nouns and verbs as in EK100. Therefore, we adapt our model to directly provide  predictions for each action class. We train our model for 446 (out of 453) classes, as we removed some videos from the dataset due to the issues with the metadata. 
% We leverage these egocentric benchmarks to showcase the effectiveness of our approach.

\subsection{Implementation Details}
\noindent\textbf{Features.} For EK100, we experiment with TBN~\cite{kazakos2019epic}, SlowFast visual~\cite{feichtenhofer2019slowfast}, and auditory~\cite{kazakos2021slow} features. We observe that using SlowFast features shows superior performance than TBN. Thus, we report all experiments using SlowFast features. We provide TBN experiments in the \textbf{supplementary}.
We extract features at 5 FPS for training the proposal generator, and we max-pool them temporally for the proposal classification part. SlowFast features have dimensionality of $D$ = 2304. For EK100, both backbones are pre-trained on EK100 recognition task.
For HOMAGE, the auditory SlowFast is pre-trained on VGG-Sound \cite{chen2020vggsound}, and the visual on EK100.

\vspace{3pt}
\noindent\textbf{Proposal generation.} We use BMN~\cite{lin2019bmn} as our $\mathcal{G}$. In~\cite{lin2019bmn} the input is rescaled to a fixed size before being fed to the network. Given that the datasets are dense and contain mostly  atomic actions, we implemented the sliding window approach (similarly to~\cite{qing2021stronger}). We use the sliding window of size 256 and a stride of 128 (160 and 80 for HOMAGE). We show the increase in average recall when using the sliding windows compared to the rescaling the inputs, as well as the ablation for the best window size in the \textbf{supplementary}. We find a simple concatenation of visual and audio features, followed by a FC layer, to be an effective strategy to fuse the modalities (early fusion). We apply Soft-NMS~\cite{bodla2017soft} as post-processing. 

\begin{table*}[!b]
    \centering
    \small
    \caption{\textbf{Showing how uni-modal and multi-modal inputs affect the performance on EK100, measured by the average mAP.} \textit{A},\textit{ V}, and \textit{ AV }- auditory, visual, and audiovisual inputs, respectively (e.g. \textit{$\mathcal{G}$-V} x \textit{$\mathcal{C}$-AV} means that we input video features to proposal generator and audiovisual to the proposal classifier). We report results of the validation set.  }
    \small
    % \resizebox{\columnwidth}{!}{%
    \begin{tabular}{c|ccc|ccc|ccc}
            \toprule
          & \multicolumn{3}{c|}{ Noun } &  \multicolumn{3}{c|}{ Verb  } &  \multicolumn{3}{c}{ Action }   \\
          \hline
        
          & $\mathcal{C}$-A  & $\mathcal{C}$-V & $\mathcal{C}$-AV & $\mathcal{C}$-A  & $\mathcal{C}$-V & $\mathcal{C}$-AV & $\mathcal{C}$-A  & $\mathcal{C}$-V & $\mathcal{C}$-AV \\
          \hline
        $\mathcal{G}$-A& 2.00 & 9.01 &  9.81 &
              2.00 & 8.17 & 08.94 & 
              0.45 & 5.65 & 6.70  \\
        $\mathcal{G}$-V&  1.60 & 10.64 & 12.48  & 
               1.76 & 10.59 & \textbf{11.96} & 
               0.59  & 7.06 &  7.66  \\
        $\mathcal{G}$-AV&  2.03 & 11.22 &\textbf{12.63} &  
               2.10 & 10.01 & 11.47 &  
               0.71 & 7.69 & \textbf{8.35} \\

    \bottomrule
    \end{tabular}%
    % }
\label{tab:bmnxclassifier}
\end{table*}

\vspace{3pt}
\noindent\textbf{Proposal classification.}
In OWL both the transformer encoder and decoder have 1 layer and 8 attention heads with the hidden unit dimension of 512. We experiment using learned and fixed positional encodings, and find that the learned perform better. The dimensionality of positional encodings $D^e = 32$. We also provide  baselines with various mutlimodal fusion strategies in the \textbf{supplementary}. These baselines perform  worse than OWL.

\subsection{Quantitative Results}
\label{sec:quantitative}
\noindent{\bf Audiovisual  impact.} 
Before incorporating context with OWL, we validate a simple baseline to verify the impact of the auditory signal on $\mathcal{G}$ and $\mathcal{C}$.
 Here, instead of using the transformer, we simply concatenate audiovisual inputs and use FC layer to encode the proposal feature.
 \begin{wraptable}{r}{8.5cm}
\captionsetup[subfloat]{font={scriptsize, color=bmv@captioncolor},labelfont=scriptsize}
\centering
\caption{\textbf{The effect of attention window size W} on in the transformer block described in Sec.~\ref{subsec:context}. We report the performance on the validation set, measured by the average mAP ($\%$). Each token on the attention pattern can attend to $\frac{W}{2}$ tokens from each side.}
\begin{subtable}{\linewidth}
\tabcolsep=0.1cm
\centering
\caption{EK100}
\small
\begin{tabular}{r|c|c|c|c|c|c|c|c}
     \toprule
     W      & 0     & 4     & 16    & 32    & 64             & 128            & 256   & 512   \\
     \toprule
     Noun   & 12.52 & 13.33 & 13.32 & 13.22 & \textbf{13.96} & 13.89          & 13.23 & 12.64 \\
     Verb   & 11.86 & 11.60 & 11.39 & 12.15 & 11.67          & \textbf{12.16} & 11.64 & 11.53 \\
     Action & 8.21  & 8.71  & 8.90  & 9.06  & \textbf{9.29}  & 8.78           & 8.58  & 8.66  \\
    \bottomrule

\end{tabular}
\label{tab:context_size}
\end{subtable}

\begin{subtable}{\linewidth}
\tabcolsep=0.1cm
\centering
\caption{Homage}
\small
\begin{tabular}{r|c|c|c|c|c|c|c|c}
     \toprule
     W      &  0   &   2  &     4         &   5   &   6    &      7   &     8    &    9  \\
     \toprule
     Action & 8.17 & 9.11 & \textbf{9.59} & 9.43  &  9.46  &    9.07  &   8.78   &   8.64 \\
    \bottomrule

\end{tabular}
\label{tab:homage_context_size}
\end{subtable}
% \vspace{-3mm}
\end{wraptable} 
 We demonstrate the performance for 9 combinations of inputs in  Tab.~\ref{tab:bmnxclassifier}: $\mathcal{G}$ with visual (V) and/or auditory (A) inputs followed by $\mathcal{C}$ with visual (V) and/or auditory (A) inputs. We find that the audiovisual classifier ($\mathcal{C}$-AV) achieves the best results for all tasks (noun, verb, action). Furthermore, the audiovisual generator ($\mathcal{G}$-AV) performs the best for noun and action. This finding validates our intuition that audio is a complementary signal to the video for detecting egocentric actions for both localization and recognition. Our hypothesis is that audio helps localize actions in situations where visual interactions are occluded (an obstacle, bad camera view), unclear (dark environments), or ambiguous, and where the audio signal is strong enough and discriminative. We discuss these scenarios in Sec.~\ref{subsec:qual}. Note that our naive audiovisual baseline ($8.35\%$) already improves the action mAP by 1.3\%, when comparing to visual-only performance ($7.06\%$). We will further refer to the visual-only model as VM. 

\begin{wraptable}{r}{4.1cm}
% \vspace{-12pt}
% \vspace{-3mm}
\centering
\small
\caption{\textbf{Action localization on HOMAGE.} We compare the performance of visual-only model (VM) \vs OWL.}
\begin{tabular}{l|cc}
\toprule
Method  & VM  & OWL  \\
\midrule
Average mAP  & 6.16 & 9.51 \\
\bottomrule
\end{tabular}
% \vspace{-12pt}
\label{tab:homage}
\end{wraptable}
\vspace{5pt}
\noindent{\bf Incorporating context.}
In Tab.~\ref{tab:context_size}, we ablate on the attention window size $W$. We find that increasing the window size \textit{does} improve the performance of our model, validating our theory that the temporal context is useful for the proposal classification. Specifically, for EK100 $W$ = 32 ($9.06\%$) and $W$ = 64 ($9.29\%$) give us the best action average mAP. Using smaller window performs comparable to the audiovisual baseline. Enlarging the window further, degrades the performance slightly, suggesting that temporally distant proposals become irrelevant. Similarly, for HOMAGE increasing $W$ improved the performance and reached its peak of $9.59\%$ with $W = 4$. Recall, that EK100 has on average $\sim$8.6 times more instances per video. Overall, our findings are similar to \cite{kazakos2021little}. However, \cite{kazakos2021little} measures the context window size in actions and OWL in proposals. As proposals are more dense, noisy, and can be classified as background, our optimal window size is larger.

\begin{table}[!h]
    \captionsetup[subfloat]{font={scriptsize, color=bmv@captioncolor},labelfont=scriptsize}
    \centering
    \caption{\textbf{Action localization on EK100.} We measure mAP@tIoU for tIoU $\in\{0.1, 0.2, 0.3,$ $0.4, 0.5\}$ and the average mAP on the validation and test sets. For reporting results on the test set, we \textbf{do not use} validation set for training, compared to~\cite{damen2021rescaling}. }
    % Second column indicates feature backbones used for the ablation: TSN~\cite{TSN2016ECCV}, I3D~\cite{carreira2017quo}, SF(A)~\cite{kazakos2021slow}, SF(V)~\cite{feichtenhofer2019slowfast}.
    % \small
	\begin{subtable}{\linewidth}
	    \centering
        \tabcolsep=0.15cm
        \resizebox{\columnwidth}{!}{
        	\begin{tabular}{r|cccccc|cccccc}
            \toprule
            Method & \multicolumn{6}{p{6.0cm}|}{\centering mAP (Val) for Noun classes @tIoU} &
                     \multicolumn{6}{p{6.0cm}}{\centering mAP (Test) for Noun classes @tIoU}  \\
            &  0.1  &  0.2  &  0.3 &  0.4 &  0.5  & Avg. &
                 0.1  &  0.2  &  0.3 &  0.4 &  0.5  & Avg. \\
            \midrule
            Damen \etal~\cite{damen2021rescaling}  &
                    10.31 & 8.33 & 6.17 & 4.47 & 3.35 & 6.53 &
                    11.99 & 8.49 & 06.04 & 4.10 & 2.80 & 6.68 \\
            AGT~\cite{nawhal2021activity}  &
                    11.63 & 9.33 & 7.05 & 6.57 & 3.89 & 7.70 & 
                    - & - & - & - & - & - \\
            \textbf{OWL} (ours)  &
                    \textbf{17.94 }& \textbf{15.81} & \textbf{14.14} & \textbf{12.13} & \textbf{9.80} & \textbf{13.96} &
                    \textbf{16.78} & \textbf{15.22} & \textbf{13.60} &\textbf{ 11.64} & \textbf{9.74} & \textbf{13.40} \\
            \bottomrule
        	\end{tabular}
        }
        	\caption{Noun} 
	\end{subtable}
	\\
	
	\begin{subtable}{\linewidth}
	    \centering
	    \tabcolsep=0.15cm
	    \resizebox{\columnwidth}{!}{%
    	\begin{tabular}{r|cccccc|cccccc}
        \toprule
        Method &\multicolumn{6}{p{6.0cm}|}{\centering mAP (Val) for Verb classes @tIoU} &
                 \multicolumn{6}{p{6.0cm}}{\centering mAP (Test) for Verb classes @tIoU}  \\
        &  0.1  &  0.2  &  0.3 &  0.4 &  0.5  & Avg. &
          0.1  &  0.2  &  0.3 &  0.4 &  0.5  & Avg. \\
        \midrule
        Damen \etal~\cite{damen2021rescaling} &
                10.83 & 9.84 & 8.43 & 7.11 & 5.58 & 8.36 &
                11.10 & 9.40 & 7.44 & 5.69 & 4.09 & 7.54  \\
        
        AGT~\cite{nawhal2021activity} &
                12.01 & 10.25 & 8.15 & 7.12 & 6.14 & 8.73  & 
                - & - & - & - & - & - \\
        \textbf{OWL} (ours)  &
                \textbf{14.48} & \textbf{13.05} & \textbf{11.82} & \textbf{10.25} & \textbf{8.73} & \textbf{11.67} &
                \textbf{16.78 }& \textbf{15.43} & \textbf{14.01} & \textbf{12.73} & \textbf{11.24} & \textbf{14.04} \\
        \bottomrule
    	\end{tabular}%
    }
    \caption{Verb}
	\end{subtable}
    \\

	\begin{subtable}{\linewidth}
	    \centering
	    \tabcolsep=0.15cm
	    \resizebox{\columnwidth}{!}{%
    	\begin{tabular}{r|cccccc|cccccc}
        \toprule
        Method & \multicolumn{6}{p{6.0cm}|}{\centering mAP (Val) for Action classes @tIoU} &
                 \multicolumn{6}{p{6.0cm}}{\centering mAP (Test) for Action classes @tIoU}  \\
         & 0.1  &  0.2  &  0.3 &  0.4 &  0.5  & Avg. &
          0.1  &  0.2  &  0.3 &  0.4 &  0.5  & Avg. \\
        \midrule
        Damen \etal~\cite{damen2021rescaling}  &
                6.95 & 6.10 & 5.22 & 4.36 & 3.43 & 5.21 &
                6.40 & 5.37 & 4.41 & 3.36 & 2.47 & 4.40 \\
        
        AGT~\cite{nawhal2021activity}  &
                7.78 & 6.92 & 5.53 & 4.22 & 3.86 & 5.66 &
                - & - & - & - & - & - \\
        \textbf{OWL} (ours) &
                \textbf{11.01} & \textbf{10.37} & \textbf{9.47} & \textbf{8.24} & \textbf{7.26} & \textbf{9.29} &
                \textbf{9.69} & \textbf{9.03} & \textbf{8.07} & \textbf{7.11} & \textbf{6.23} & \textbf{8.03} \\

        \bottomrule
    	\end{tabular}%
    }
    	\caption{Action}
	\end{subtable}
	\label{tab:sota}
% \vspace{-3mm}

\end{table}
\vspace{10pt}
\noindent{\bf Comparison with the state-of-the-art.}
We compare the performance of OWL on EK100 with the existing methods in Tab.~\ref{tab:sota}. OWL performs significantly better than the baseline of~\cite{damen2021rescaling}, and~\cite{nawhal2021activity}, and achieves $9.29\%$ average mAP for the action class. For HOMAGE, to the best of our knowledge, we are the first work to explore it for TAL. As shown in Tab.~\ref{tab:homage}, OWL achieves $9.5\%$ average mAP, which is decent performance for more diverse dataset activities and a good baseline score to encourage more contributions from future work. In addition, we compare OWL with VM to validate the effectiveness of our approach to incorporate audio. OWL significantly outperforms VM by $3.35\%$ average mAP.

\subsection{Performance Analysis and Qualitative Results}
\label{sec:experiments-analysis}
\noindent\textbf{Visual occlusion analysis.} We validate the hypothesis that OWL helps to detect actions in visually occluded environments by comparing the mAP of \textit{more-occluded} \vs \textit{less-occluded} instances. To define the occlusion level, we assume that the visual occlusion must happen in the place of hand-object interactions. We utilize the detected hand-object interactions in EK100~\cite{Shan_2020_CVPR}. We measure the percentage of occluded frames per action instance by considering a frame as occluded when the object's bounding box of the interaction is missing. Then, we divide the validation set into 3 disjoint partitions: \textit{No occlusion}, \textit{Low occlusion} ($<8\%$), \textit{High occlusion} ($>8\%$). We empirically find that $8\%$ of occluded frames balances the size of the three partitions. We then evaluate VM and OWL on these partitions, and measure the improvement in performance. As shown in Tab.~\ref{tab:occlusion} both models achieve the lowest performance on \textit{High occlusion} across all tasks. As we hypothesized, we achieve the highest performance boost when using OWL over VM on \textit{High occlusion} instances ($56.7\%$ improvement on action task \vs only $18.0\%$ with \textit{Low occlusion} and $22.2\%$ with \textit{No occlusion}). 
\newline
\begin{table}[ht]
\captionsetup[subfloat]{font=scriptsize,labelfont=scriptsize}
\centering
\caption{\textbf{Visual occlusion analysis.} 
We breakdown the performance of VM and OWL on EK100 into 3 partitions: \textit{No, Low, and High} occlusion, based on the percentage of missing predictions of the hand-objects interactions \cite{Shan_2020_CVPR,damen2021rescaling}. Intuitively, when the object is out of the frame (occluded), the hand-object interactions are missing. OWL improves the performance across the board, especially in \textit{High occlusion} subset.
}

\resizebox{\columnwidth}{!}{
    \begin{tabular}{c|ccc|ccc|ccc|ccc}
    \toprule
    % \multirow{3}{}{} & \multicolumn{9}{|c|}{Occlusion level} & \multicolumn{3}{c}{Validation set} &
    %  & \multicolumn{3}{c|}{No occlusion} & \multicolumn{3}{c|}{Low occlusion} & \multicolumn{3}{c|}{High occlusion} & \multicolumn{3}{}{} \\
    \multicolumn{1}{l}{} & \multicolumn{3}{|c|}{No occlusion} & \multicolumn{3}{c|}{Low occlusion} & \multicolumn{3}{c|}{High occlusion} & \multicolumn{3}{c}{Validation set} \\

    \hline
    \multicolumn{1}{l|}{} & noun          & verb          & action        & noun          & verb          & action        & noun          & verb          & action        & noun          & verb          & action        \\
    % \hline
    VM mAP       & 16.0          & 14.8          & 10.8          & 14.3          & 16.0          & 12.8          & 9.4           & 10.0          & 6.0           & 10.6          & 10.6          & 7.1           \\
    % \hline
    OWL mAP              & 19.4          & 16.4          & 13.2          & 17.7          & 20.0          & 15.1          & 12.9          & 14.5          & 9.4           & 14.0          & 11.7          & 9.3           \\
    % \hline
    Improvement  in \%   & \textbf{21.3} & \textbf{10.8} & \textbf{22.2} & \textbf{23.8} & \textbf{25.0} & \textbf{18.0} & \textbf{37.2} & \textbf{45.0} & \textbf{56.7} & \textbf{31.2} & \textbf{10.2} & \textbf{31.6} \\
    \hline
    \# instances         & \multicolumn{3}{c|}{4879}         & \multicolumn{3}{c|}{2407}          & \multicolumn{3}{c|}{2382}           & \multicolumn{3}{c}{9668}     \\
    \bottomrule
    \end{tabular}
}
\label{tab:occlusion}
% \vspace{-5mm}
\end{table}

\begin{figure*}[h]
\begin{center}
\includegraphics[width=\textwidth]{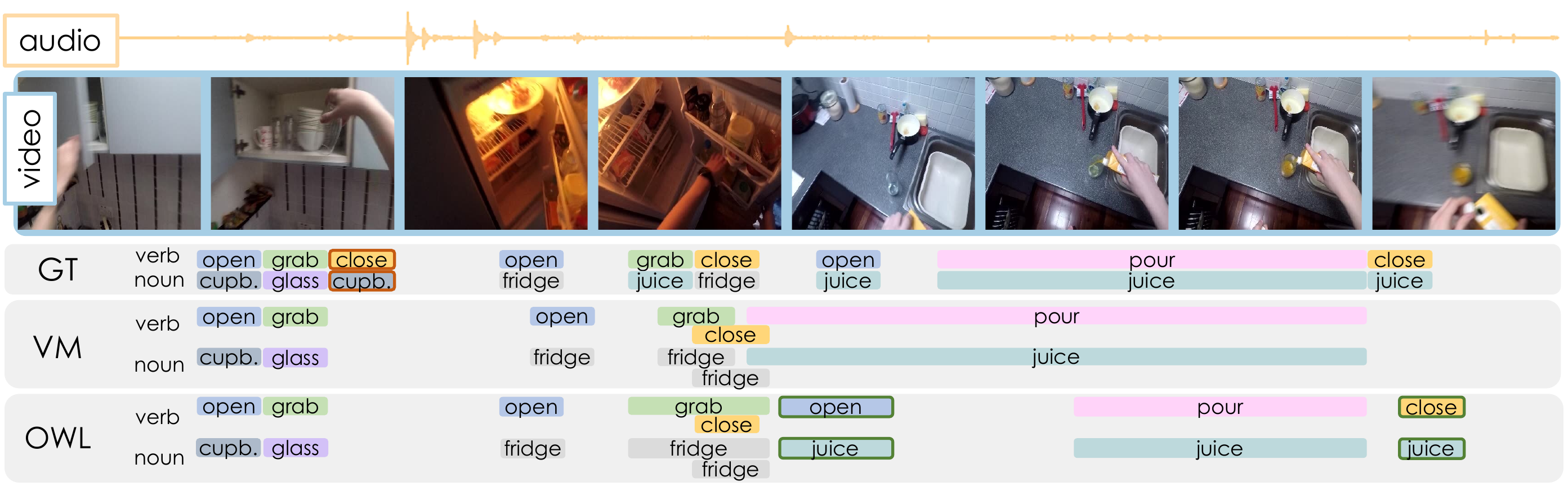}
\small
\caption{\textbf{Qualitative results.} The same scenario as in Fig.\ref{subfig:context}. The ground truth (GT) annotations are compared with the predictions of VM and OWL. We highlight with \textcolor{red}{red} the GT actions that were missed by both VM andd OWL, and with \textcolor{green}{green} the predictions where only either of them succeeds. We can observe how OWL produces temporally more precise and complete outputs.}
\label{figure:qual}
\end{center}
\vspace{-7mm}
\end{figure*}

\noindent\textbf{Qualitative Results.} Fig.~\ref{figure:qual} visualizes the localization results of OWL  and compares it to the results of VM. As we can see, VM fails to predict \textit{open juice} and \textit{close juice}. However, OWL predicts them successfully. Furthermore, the localized actions are on average more precise for OWL (\textit{open fridge}, \textit{pour juice}). Our intuition is that the pouring sound helps to localize the actions better.

\label{subsec:qual}

\section{Limitations}
\label{sec:lims}
The scope of this work is limited to audio-visual context for TAL in untrimmed unedited videos with a large number of action categories per video.
We acknowledge that the audio and visual signals may have an interesting interplay in highly edited videos, \eg those on YouTube, TikTok, and movies. However, the audio signal in edited video might not be predominantly associated with the action. It might also be mixed with speech and music to evoke emotions in the viewers. Thus, we believe that it is healthy to explore the two lines of research independently, edited \vs unedited. 
\section{Conclusion}
\label{sec:conclusion}
This work studies multi-modal TAL using egocentric unedited videos.
The specific challenges of public egocentric video benchmarks (\eg, unedited footage, localization of actions out of frame and large number of action classes per video) invite to rethink the inner workings of the TAL models. This work does so by means of two booster principles: multi-modality, with audio, and temporal continuity to complement the visual signal.
We validate our hypothesis by experimenting with the multiple audiovisual fusion approaches as well as context-aware pipelines. A technical contribution of our work is OWL, a transformer-based model that leverages both temporal context and modality fusion. By using OWL, we achieve competative performance on EK100, and make a strong baseline record on HOMAGE. 

\noindent{\bf Acknowledgements.}
\textbf{V.E. contributions} involved conceptualisation, methodology, writing - review and editing, and supervision. This work was supported by the King Abdullah University of Science and Technology (KAUST) Office of Sponsored Research through the Visual Computing Center (VCC) funding.

\bibliography{egbib}

\clearpage
\appendix
%%%%%%%%%%%%%%%%%%%%%%%%%%%%%%%%%%%%%%%%%%%%%%
\section*{Supplementary Material}
\noindent 

We complement our work with the following: (i) The details on the proposal generation (Sec.~\ref{sec:supp_proposals}), (ii) per-class performance analysis (Sec.~\ref{sec:supp_class_analysis}),  (iii) fusion experiments (Sec.~\ref{sec:fusion}), and (iv) qualitative examples \textbf{(please check the attached slides).}

\section{Action proposals}
\label{sec:supp_proposals}
This section analyzes the action proposals for EPIC-Kitchens-100 (EK100) produced by the proposal generator, as explained in Sec.~\textcolor{red}{3.1} and Fig.~\textcolor{red}{2} (\cf the main manuscript). We measure the quality of the proposals with average recall (AR).  \cite{escorcia2016daps}
It is worth noting that proposals are class-agnostic and require further classification. AR measures the localization quality of the action proposals. We consider the limited number of predicted proposals when computing AR and compute it for several tIOU thresholds.  In the following sections, we investigate which feature encoders to use and how to treat the input sequence.
\begin{wraptable}{r}{6.4cm}
    \centering
    \begin{tabular}{c|c| c}
    \toprule
    Features &  Modality &  {AR ($\%$)} \\
      \midrule
       TBN  & RGB, flow, audio &  64.61 \\
       SlowFast & visual & 64.09 \\
       SlowFast & audio & 56.38 \\
       SlowFast & visual, audio & \textbf{65.66} \\
    \bottomrule
    \end{tabular}
    \caption{\textbf{Average Recall (AR) on EK100} for the proposals using TBN and SlowFast features in uni-modal and multi-modal scenarios.}
    % We investigate which feature encoder results in superior performance and whether the multi-modal scenario performs better than the uni-modal.
    \label{tab:recall_tbn_vs_sf}
    \vspace{-25pt}
\end{wraptable}
\subsection{Feature encoders}
Our focus is to investigate audiovisual inputs; thus, we consider the encoders that process auditory and visual signals. We consider TBN~\cite{kazakos2019epic} and SlowFast~\cite{feichtenhofer2019slowfast,kazakos2021slow} networks as our feature encoders. TBN operates on RGB, Flow, and spectrogram.  Visual and Auditory SlowFast take video frames and spectrogram, respectively, as inputs. In Tab.~\ref{tab:recall_tbn_vs_sf} we compare the performance of the proposal generator on EK100 with TBN and SlowFast features. To demonstrate the effect of audiovisual features, we also provide the results of a uni-modal proposal generator with visual-only or audio-only inputs. To create audiovisual SlowFast features, we concatenate visual and auditory features of the corresponding SlowFast backbones. We notice that audiovisual SlowFast features  outperforms TBN ($65.66\%$ \vs $64.61\%$). Furthermore, we can observe that multi-modal SlowFast features outperforms uni-modal ($65.66\%$ for audiovisual  \vs $64.09\%$ for visual and $56.38\%$  for audio). 
\begin{wraptable}{r}{6.5cm}
    \centering
    \begin{tabular}{c|c}
    \toprule
    Features & {AR ($\%$)} \\
    \midrule
       TBN (rescaled)  & 54.91 \\
       TBN (sliding window)  & \textbf{64.61} \\
    \bottomrule
    \end{tabular}
    \caption{\textbf{Average Recall (AR) on EK100} for the proposals treating the input sequence with rescaling \vs using the sliding windows.}
    \label{tab:recall_rescale}

\end{wraptable}

\subsection{Input sequence }
As videos can vary in duration, their features can have different temporal dimensions. We investigate two types of input sequence treatment in the proposal generator: (1) rescaling the features to produce the input of a particular temporal size and (2) iterating over the features with a sliding window. Sliding window treats time as the reference framework, whereas  feature rescaling uses duration. As  mentioned in \cite{zhao2021video}, rescaling features is suboptimal for detecting short actions in long videos. This is particularly relevant for our work as EK100 is annotated with many atomic instances, and a video duration can exceed one hour. We observed that previous approaches for the temporal action localization in the dataset used both strategies. For instance, \cite{damen2021rescaling} utilizes feature rescaling and~\cite{qing2021stronger} uses the sliding window. Tab.~\ref{tab:recall_rescale} compares the average recall (AR) of proposals using either strategy. We can see that the sliding window approach results in $10\%$ AR increase compared to rescaling. That validates the idea that the sliding window is a better way to deal with the atomic actions in the dataset.  Therefore, we conduct our experiments using the sliding window approach.
\begin{wraptable}{r}{5.0cm}
    \centering
    \begin{tabular}{c| c}
    \toprule
     Window Size &  {AR ($\%$)} \\
      \midrule
       200  & 65.52 \\
       300 &  \textbf{65.66} \\
       400 &  63.75 \\
    \bottomrule
    \end{tabular}
    \caption{\textbf{Average Recall (AR) on EK100 for the proposals using different sliding window sizes. }}
    \label{tab:recall_win}
\end{wraptable}
\subsection{Window size}
While processing the input sequence with a sliding window, we aim for the most effective window size. As  observed in \cite{qing2021stronger}, over $98\%$ of annotated action instances in EK100~\cite{damen2021rescaling} are shorter than 20 seconds. We extracted features at 5 fps; thus, to capture 98\% of actions, we should aim for a minimal stride $s = 20 \times 5  = 100$.  In our experiments, we always make the window size $w$ double the stride $s$.
In Tab.~\ref{tab:recall_win} we investigate the best window size, starting with with $w = 200$  and $s = 100$. We keep increasing $w$ and $s$  until the performance degrades. That ensures that at least one sliding window will cover any action that does not exceed $\frac{w}{2}$. We reach the highest performance with $w = 300$ (and $s = 150$). This is because increasing the window size to 300 incorporates some relevant context to the model. However, further increasing the window size to 400 degrades the performance, suggesting that faraway context becomes irrelevant (similar to OWL's temporal context).

\begin{figure*}[h]
    \captionsetup[subfloat]{font={scriptsize, color=bmv@captioncolor},labelfont=scriptsize}
    \centering
    \small
    \begin{subfigure}[t]{\linewidth}
        \centering
        \caption{Verbs classes}
         \includegraphics[width=\textwidth]{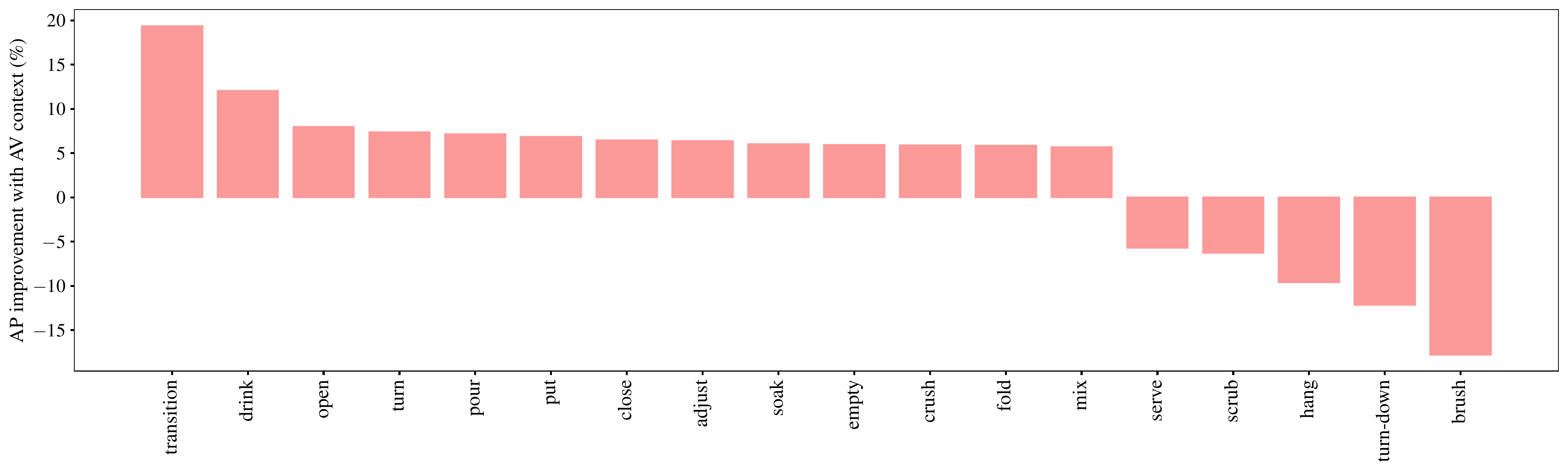}
        \label{fig:verb-aps}
    \end{subfigure}%

    \begin{subfigure}[t]{\linewidth}
        \centering
        \vspace{-10pt}
        \caption{Noun classes}
        \includegraphics[width=\linewidth]{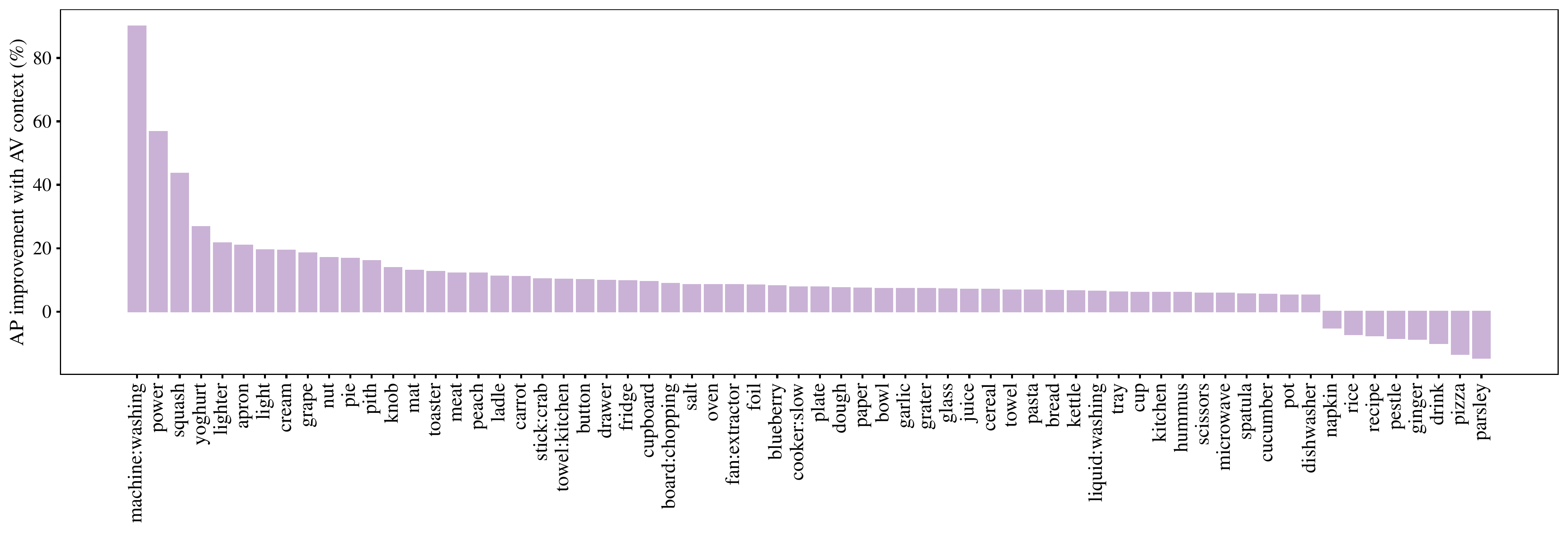}
        \label{fig:noun-aps}
    \end{subfigure}

    \caption{\textbf{Absolute per class improvement in performance of OWL with respect to VM, measured by the average precision (AP, $\%$).} To observe the significant changes, we only visualize classes with the absolute difference in AP greater than $5\%$. 
    }
    \label{fig:dataset-statistics}
\end{figure*}

\section{Per-class performance of OWL}
\label{sec:supp_class_analysis}

In Fig.~\ref{fig:dataset-statistics} we show a per-class performance comparison of OWL \vs the visual-only model (VM) on EK100. We plot the absolute improvement, measured by average precision (AP), for noun (Fig.~\ref{fig:noun-aps}) and verb (Fig.~\ref{fig:verb-aps}) classes. We can observe that OWL performs better than VM for most verb and noun classes. We attribute the improvements to audio or context incorporation and discuss them in the following subsections.

\noindent{\bf Audio.}
Verbs \textit{pour}, \textit{crush}, \textit{drink} have distinctive sounds, and OWL performs better than VM on these classes. \textit{Drink}, is an interesting case as the source of sound is very close to the camera microphone. As we expect, OWL improves by more than $10\%$ on this class. Likewise, several nouns, such as \textit{machine:washing}, \textit{microwave}, \textit{fridge}, \textit{kettle},  \textit{fan:extractor}, etc. are electronic appliances which usually have distinctive sounds when turned on/off and while operating.

\noindent{\bf Context.}
Several verbs, such as \textit{transition}(used interchangeably with \textit{move}, \textit{walk in} in the dataset taxonomy), \textit{open}, \textit{put}, \textit{close} have better predictions with OWL. We believe that the improvement for these verb classes can be attributed to context incorporation. As  mentioned in Fig.~1 of the main paper, humans often do their kitchen activities following some patterns (logical order in human-object interactions).
We also hypothesize that food that is  packed, such as \textit{grape}, \textit{nut}, \textit{meat}, \textit{carrot}, \textit{salt}, \textit{juice}, \textit{cereal}, \textit{pasta}, etc. could be ambiguous for the model when shown packed. 
\section{Fusing audio and visual modalities}
In this section we explain our preliminary experiments on the multi-modal fusion strategies.
First, we elaborate on our terminology of the proposal generator and classifier.
\label{sec:fusion}

\noindent\textbf{Proposal generator $\mathcal{G}$.}  Given the visual features $\mathbf{x}^v$ and the audio features $\mathbf{x}^a$ of the video sequence, the proposal generator $\mathcal{G}$ predicts a set of candidate  segments with temporal boundaries, namely, proposals $\Phi  = \left\{\phi_m =\left({t}_{\textrm{s},m},{t}_{\textrm{e},m}, {s}_m \right) \right\}_{m=1}^{M}$,  where $\phi_m$ represents an action proposal, $M$ is the number of proposals, and ${t}_{\textrm{s},m}$, ${t}_{\textrm{e},m}$and ${s}_m$ are its start time, end time and confidence score, respectively. Note that proposals do not have class labels.

\noindent\textbf{Proposal classifier $\mathcal{C}$.} Given the set of proposals $\Phi$, the snippet-level visual features $\mathbf{x}^v$, and audio features $\mathbf{x}^a$, we first extract visual features $\mathbf{x}_m^v$  and audio features $\mathbf{x}_m^a$ for the $m^{\textrm{th}}$ proposal  by max-pooling  the snippets  within its start/end boundaries\footnote{We round the start/end values to the nearest snippets indices.}. Then, the proposal classifier $\mathcal{C}$ predicts from  $\mathbf{x}_m^v$ and $\mathbf{x}_m^a$  verb  and noun class labels $c^{\textrm{verb}}$ and $c^\textrm{noun}$, as well as their respective scores $s^\textrm{verb}$ and $s^\textrm{noun}$. Based on the predicted verbs and nouns, we generate action predictions $\Psi  = \left \{ \psi_n=\left ({t}_{\textrm{s},n}, {t}_{\textrm{e},n}, {c}_n, {s}_n\right )  \right \}_{n=1}^{N}$ .
, where   $c_n = (c^\textrm{verb}_i, c^\textrm{noun}_j) \in \mathcal{A}$ and  $s_n = s_i^\textrm{verb} s_j^\textrm{noun}$. $\mathcal{A}$ is a set of pre-defined actions, each composed of a noun and a verb, and  $1 \leq i \leq M$ and $1 \leq j \leq M$ are proposal indices.

% \subsection{Which modality to use?}
% The first question is how helpful the visual and audio modalities are to each temporal action localization pipeline stage. 
%  For each stage, we can either: use only visual features (as most existing methods), use only the audio features, or use both.  As we will discuss in  Sec.~\ref{sec:quantitative}, for both \textit{$\mathcal{G}$} and \textit{$\mathcal{C}$} using both visual and audio modalities is the most effective strategy.  In the following sections, \textit{we focus on exploring $\mathcal{C}$.} Particularly, we will compare and study different strategies to fuse the modalities, discussing where and how to fuse them.
 \begin{figure*}[ht]
\begin{center}
\includegraphics[width=\textwidth,clip]{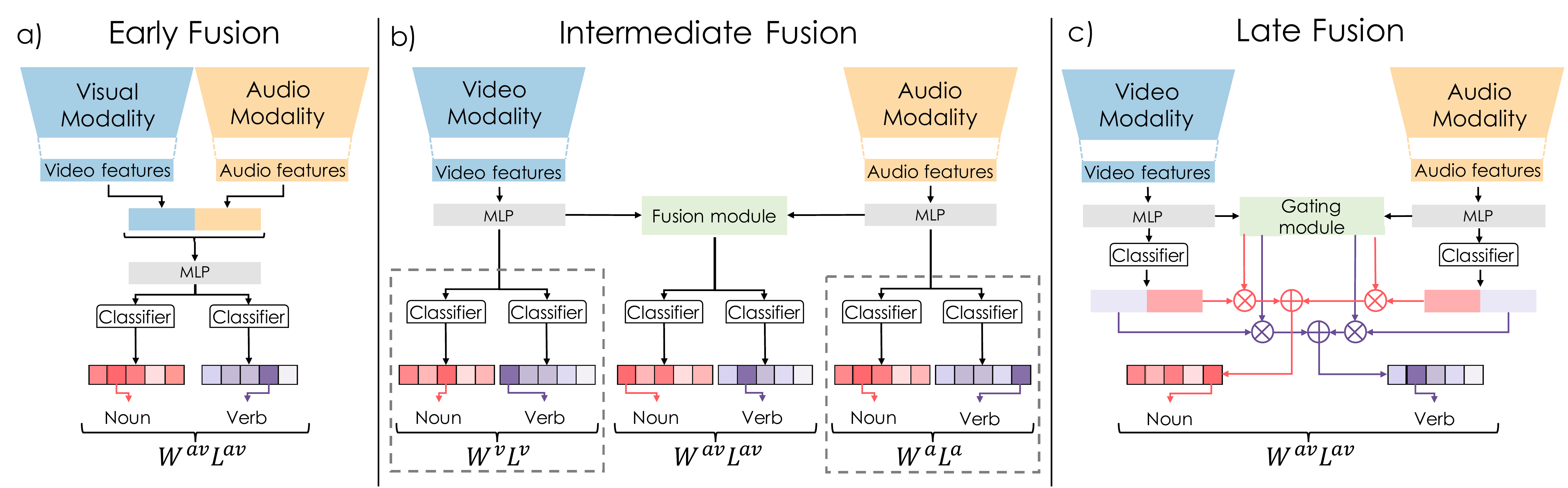}
\end{center}
\small
\caption{\textbf{Fusion methods for the audio and video streams.} Early fusion (a) does features aggregation. Intermediate fusion (b) combines intermediate representations of each modality. The model can be trained jointly by optimizing for three losses, or we can simplify it by setting $W_v = W_a = 0$ (the affected branches are highlighted with dashed lines). Late fusion (c) combines scores of two modalities. The gating module produces per-class weight for the scores generated by each modality. The weighted scores are aggregated by summation. Here we illustrate \textit{cross-gating}, in which the gating module takes representations of both modalities as the input.}
\label{figure:fusion}
\end{figure*}

\subsection{Where and how to fuse the modalities in $\mathcal{C}$?}
 We categorize the modality fusion into the following: early, late, and intermediate fusion, as shown in Fig.~\ref{figure:fusion}. 

\textit{Early fusion} happens at the input feature level  (Fig.~\ref{figure:fusion}~a). Given the proposal's visual features $\mathbf{x}_{m}^{v}$ and audio features $\mathbf{x}_{m}^{a}$, we first fuse them and obtain one singe feature vector  $\mathbf{x}_{m} = \mathcal{F}_{\textrm{early}} (\mathbf{x}_{m}^{v}, \mathbf{x}_{m}^{a})$. We feed $\mathbf{x}_{m}$ to the following layers of operations (e.g., MLP), and classify it into different noun and verb classes. \textit{How to choose the fusing function $F_{\textrm{early}}$?} In our analysis, we simply fuse the modalities by concatenating the visual and audio features along the channel dimension. This doesn't require extra computations and counts on the following network layers to learn from the fused features.

\textit{Intermediate fusion} happens at the intermediate feature level (Fig.~\ref{figure:fusion}~b). We process the audio and video features independently for certain layers, and generate intermediate features $(\mathbf{z}_m^v$ and $\mathbf{z}_m^a)$. We fuse them to one feature via $\mathbf{z}_m = \mathcal{F}_{\textrm{inter}}(\mathbf{z}_m^v,\mathbf{z}_m^a)$. The fused features $\mathbf{z}_m$ as well as the visual and audio intermediate features $\mathbf{z}_m^v$ and $\mathbf{z}_m^a$ are processed independently in the following layers, and correspondingly predict three groups of classification scores. We use them all for training, and only use the scores from the fused features for inference. Similarly to early fusion, we use concatenation for  $\mathcal{F}_{\textrm{inter}}$ in our experiments (Tab.~\ref{tab:fusion_results_all}). Our proposed model OWL uses intermediate fusion; however, instead of concatenation, it adaptively fuses audio features to visual by correlating to the context (more in Sec.~\ref{sec:OWL}).

\textit{Late fusion} happens at the output score level (Fig.~\ref{figure:fusion}~c). The visual and audio features of all proposals are independently processed until they produce classification scores $\mathbf{s}_m^v= \{\mathbf{s}^{\textrm{verb},v}_m\in \mathbb{R}^{V}, \mathbf{s}^{\textrm{noun},v}_m \in \mathbb{R}^{U}\}$, and $\mathbf{s}_m^a= \{\mathbf{s}^{\textrm{verb},a}_m\in \mathbb{R}^{V}, \mathbf{s}^{\textrm{noun},a}_m \in \mathbb{R}^{U}\}$ where $V$ and $U$ are the numbers of verb and noun classes, respectively. We fuse the  scores from both modalities via $\mathbf{s}_m = \mathcal{F}_{\textrm{late}}(\mathbf{s}^{{v}}_m, \mathbf{s}^{{a}}_m)$, and apply \textit{softmax} to $\mathbf{s}_m$ to generate the final prediction for nouns and verbs. 
% $\mathcal{F}_{\textrm{late}}$. 
For late fusion, there is no straightforward way to do concatenation. Naively averaging or multiplying corresponding scores of the two modalities is not effective, due to the imbalance between the modalities. While audio can be a complementary source of information, it doesn't contribute equally as the visual modality to solving the task. We observe that either modality `specializes' in different classes, and it's beneficial to combine the scores with different weights per class. For example, the action of `taking something' is usually not evident from the sound, but `turning on' a kitchen device is. 

For effective {late fusion}, motivated by~\cite{miech2017learnable,DBLP:conf/bmvc/LiuANZ19}, we design a gating module to weight the per-class scores before fusing them. The gating module ${\Theta}$ is composed of  a fully-connected layer followed by a sigmoid activation function. It learns from the concatenated intermediate features of the two modalities $\mathbf{z}_m = [\mathbf{z}^{v}_m;\mathbf{z}^{a}_m]$ to predict  weights  for the verb and noun classes for both modalities:  $\mathbf{w}^v_m = \Theta^v(\mathbf{z})$, $\mathbf{w}^a_m = \Theta^v(\mathbf{a})$. The weights are applied to the classification scores $\mathbf{s}_{m}^{v} $ and $\mathbf{s}_{m}^{a}$ of two modalities for linear combination, and generate the final scores via  $\mathbf{s}_m = \mathbf{s}^{v}_m \odot \mathbf{w}^v_m + \mathbf{s}^{a}_m \odot \mathbf{w}^a_m$. We call the gating strategy \textit{cross-gating}. Alternatively, we also experiment with a \textit{self-gating} strategy, where the weights for each modality is learned only from its own features: $\mathbf{w}^v_m = \Theta^v(\mathbf{z}^v), \mathbf{w}^a_m = \Theta^a(\mathbf{z}^a)$.

\subsection{Results}
We compare several fusion strategies in Tab.~\ref{tab:fusion_results_all}.  All experiments were run on audiovisual proposals ($\mathcal{G}$-AV). Early fusion results in a significant improvement over the visual-only model (VM). The intermediate fusion with only audiovisual supervision ($8.24\%$ action mAP) does not perform better than early fusion. However, we can achieve better results by jointly training with the supervision from the visual and audio streams ($8.75\%$).  Doing late fusion with self-gating weights does not perform well (only $7.99\%$), but late fusion with cross-gating (Late F CG) achieves $8.82\%$. This finding is expected as cross-gating has richer representations of both modalities for weighting the class scores.
\begin{table}[ht!]
\centering
\caption{\textbf{Fusion methods performance on EK100, measured by the average mAP ($\%$).} SG and CG correspond to the self-gating and cross-gating scenarios described in Sec.~\ref{subsec:fusion}, respectively. We also show the modality streams being supervised in the second column. }
\small
    \begin{tabular}{r|c|c|c|c}
        \toprule
Method & Supervision & Noun & Verb & Action \\
\midrule
Early F & AV &  12.63 & 11.47 & 8.35 \\
Intermediate F & AV & 12.55 & 11.66 & 8.24  \\
Intermediate F & V, A, AV & \textbf{13.66} & \textbf{12.90} &  8.75 \\
Late F SG & V, A & 11.51 & 10.84 & 7.99  \\
Late F CG & V, A & 12.66  & 12.89 & \textbf{8.82} \\
\bottomrule
\end{tabular}
\label{tab:fusion_results_all}
\end{table}

\end{document}